\documentclass[runningheads]{llncs}
\usepackage{graphicx}
\usepackage{comment}
\usepackage{amsmath,amssymb,amsfonts} % define this before the line numbering.
\usepackage{color}
\usepackage{multirow}

\usepackage{svg}

\usepackage[accsupp]{axessibility}  % Improves PDF readability for those with disabilities.

\usepackage{booktabs} % For thickline in tables

\definecolor{lightgray}{gray}{0.9}

\usepackage{tikz}
\usetikzlibrary{shapes}
\usetikzlibrary{plotmarks}
\usepackage{ifthen}

\usepackage{pgfplots}
\pgfplotsset{width=10cm,compat=1.15}
\usepgfplotslibrary{statistics}
\usepgfplotslibrary{colormaps}
\usepgfplotslibrary{fillbetween}

\usepackage{url}
\usepackage[hidelinks]{hyperref}
\hypersetup{
    colorlinks=true,
    linkcolor=black,
    filecolor=black,
    urlcolor=gray,
    citecolor=black,
    breaklinks=true
}

\usepackage{orcidlink}

\definecolor{c2}{RGB}{ 255  0   0}
\definecolor{c3}{RGB}{          0  202.2405  255.0000}
\definecolor{c5}{RGB}{ 246.2025  131.8860    8.7975}    % SetDesc
\definecolor{c8}{RGB}{   128.0100  128.0100  128.0100}
\definecolor{c6}{RGB}{          0         0  255.0000}
\definecolor{c4}{RGB}{          0   87.9240         0}
\definecolor{c7}{RGB}{229, 194, 36}  % ST->FT
\definecolor{c1}{RGB}{19,219,30}  % AT->FT
\definecolor{c9}{RGB}{131.8966  131.8966  255.0000}

\definecolor{dc1}{RGB}{230, 25, 75} % Red
\definecolor{dc2}{RGB}{60, 180, 75} % Green
\definecolor{dc3}{RGB}{255, 225, 25} % Yellow
\definecolor{dc4}{RGB}{0, 130, 200} % Blue
\definecolor{dc5}{RGB}{245, 130, 48} %Orange
\definecolor{dc6}{RGB}{145, 30, 180} % Purple
\definecolor{dc7}{RGB}{70, 240, 240} % Cyan
\definecolor{dc8}{RGB}{240, 50, 230} % Magenta
\definecolor{dc9}{RGB}{210, 245, 60} % Lime
\definecolor{dc10}{RGB}{250, 190, 212} % Pink
\definecolor{dc11}{RGB}{0, 128, 128} % Teal
\definecolor{dc12}{RGB}{220, 190, 255} % Lavender
\definecolor{dc13}{RGB}{170, 110, 40} % Brown
\definecolor{dc14}{RGB}{255, 250, 200} % Beige
\definecolor{dc15}{RGB}{128, 0, 0} % Maroon
\definecolor{dc16}{RGB}{170, 255, 195} % Mint
\definecolor{dc17}{RGB}{128, 128, 0} % Olive
\definecolor{dc18}{RGB}{255, 215, 180} % Apricot
\definecolor{dc19}{RGB}{0, 0, 128} % Navy
\definecolor{dc20}{RGB}{128, 128, 128} % Grey
\definecolor{dc21}{RGB}{255, 255, 255} % White
\definecolor{dc22}{RGB}{0, 0, 0} % Black

\newcommand{\mtrx}[1]{\mathbf{#1}}

\tikzset{
  chart/.style={
    legend label/.style={font={\tiny},anchor=west,align=left},
    legend box/.style={rectangle, draw, minimum size=5pt},
    axis/.style={black,semithick,->},
    axis label/.style={anchor=east,font={\tiny}},
  },
  pie chart/.style={
    chart,
    slice/.style={line cap=round, line join=round, thick,draw=white},
    pie title/.style={font=\scriptsize},
    slice type/.style 2 args={
        ##1/.style={fill=##2},
        values of ##1/.style={}
    }
  }
}

\pgfdeclarelayer{background}
\pgfdeclarelayer{foreground}
\pgfsetlayers{background,main,foreground}

\newcommand{\pie}[3][]{
    \begin{scope}[#1]
    \pgfmathsetmacro{\curA}{180}
    \pgfmathsetmacro{\radius}{1}
    \def\Centre{(0,0)}
    \node[pie title] at (90:-1.6) {#2};
    \foreach \v/\s in{#3}{
        \pgfmathsetmacro{\deltaA}{\v/100*360}
        \pgfmathsetmacro{\nextA}{\curA + \deltaA}
        \pgfmathsetmacro{\midA}{(\curA+\nextA)/2}

        \path[slice,\s] \Centre
            -- +(\curA:\radius)
            arc (\curA:\nextA:\radius)
            -- cycle;

   \pgfmathsetmacro{\ysign}{ifthenelse(mod(\midA,360)<=180,1,-1)}
   \pgfmathsetmacro{\xsign}{ifthenelse(mod(\midA-90,360)<=180,-1,1)}

   \begin{pgfonlayer}{foreground}
        \draw[-,thin] \Centre ++(\midA:0.9*\radius) -- 
                               ++(\xsign*0.1*\radius,\ysign*0.2*\radius) --                      ++(\xsign*0.1*\radius,0) 
                      node[yshift=0pt,xshift=\xsign*12pt, near end,pie values,values of \s]{$\v\%$};
   \end{pgfonlayer}

        \global\let\curA\nextA
    }
    \end{scope}
}

\begin{document}
% \linenumbers
\pagestyle{headings}
\mainmatter
\def\ECCVSubNumber{7}  % Insert your submission number here

\title{Cross-Camera View-Overlap Recognition}

\begin{comment}
\titlerunning{ECCV-22 submission ID \ECCVSubNumber} 
\authorrunning{ECCV-22 submission ID \ECCVSubNumber} 
\author{Anonymous ECCV submission}
\institute{Paper ID \ECCVSubNumber}
\end{comment}
%******************

% \begin{comment}
\titlerunning{Cross-Camera View-Overlap Recognition}
\author{Alessio Xompero\orcidlink{0000-0002-8227-8529} \and
Andrea Cavallaro\orcidlink{0000-0001-5086-7858}}
\authorrunning{A. Xompero and A. Cavallaro}
\institute{Centre for Intelligent Sensing, Queen Mary University of London, London, U.K.\\
\url{http://cis.eecs.qmul.ac.uk/} \\
\email{\{a.xompero,a.cavallaro\}@qmul.ac.uk}}
% \end{comment}
%******************
\maketitle

\begin{abstract}
We propose a decentralised view-overlap recognition framework that operates across freely moving cameras without the need of a reference 3D map. Each camera independently extracts, aggregates into a hierarchical structure, and shares feature-point descriptors over time. A view overlap is recognised by view-matching and geometric validation to discard wrongly matched views. 
The proposed framework is generic and can be used with different descriptors. We conduct the experiments on publicly available sequences as well as new sequences we collected with hand-held cameras. We show that Oriented FAST and Rotated BRIEF (ORB) features with Bags of Binary Words within the proposed framework lead to higher precision and a higher or similar accuracy compared to  NetVLAD, RootSIFT, and SuperGlue.
\end{abstract}

\section{Introduction}
\label{sec:intro}

View-overlap recognition is the task of identifying the same scene captured by a camera over time~\cite{Galvez2012TRO,Gao2018IROS_LDSO,Garcia-Fidalgo2018RAL,MurArtal2015TRO,MurArtal2014ICRA,Schlegel2018RAL,Tsintotas2019RAL} or by multiple cameras~\cite{Cieslewski2018ICRA,Forster2013IROS,Riazuelo2014RAS,Schmuck2019JFR_CCMSLAM,Zou2013TPAMI} moving in an unknown environment. The latter scenario includes immersive gaming with wearable or hand-held cameras, augmented reality, collaborative navigation, and faster  scene  reconstruction. Existing methods first identify view overlaps with features associated to the whole image (view features) and then with features around interest points (local features)~\cite{Galvez2012TRO,Garcia-Fidalgo2018RAL,MurArtal2014ICRA,Schonberger2016CVPR}. 
View features can be a direct transformation of an image or an aggregation of local features. For example, convolutional neural networks can transform an image into a view feature (e.g., NetVLAD~\cite{Arandjelovic2018TPAMI} or DeepBit~\cite{Lin2018TPAMI}) or the original image can be down-sampled and filtered to use as view feature~\cite{Riazuelo2014RAS}.
{Local features} describe a small area surrounding localised interest points with a fixed-length descriptor, resulting from computed statistics (histogram-based features)~\cite{Lowe2004IJCV}, pixel-wise comparisons (binary features)~\cite{Rublee2011ICCV,Calonder2010ECCV,Leutenegger2011ICCV}, or the output of convolutional neural networks~\cite{DeTone2018CVPRw_SuperPoint,Dusmanu2019CVPR_D2Net}. When a 3D map of the scene is available (or can be reconstructed), recognising view overlaps   involves the projection of the 3D points of the map onto a camera view to search for the closest local features. Finally, triangulation and optimisation refine the 3D map and the poses of the cameras~\cite{Camposeco2019CVPR,Cieslewski2018ICRA,Forster2013IROS,MurArtal2015TRO,Riazuelo2014RAS,Sarlin2019CVPR_HFNet,Sattler2018CVPR,Schmuck2019JFR_CCMSLAM,Schonberger2016CVPR,Zou2013TPAMI}. 

The above-mentioned methods are centralised~\cite{Forster2013IROS,MurArtal2015TRO,Schmuck2019JFR_CCMSLAM,Schonberger2016CVPR,Zou2013TPAMI} (i.e.~they process images or features in a single processing unit) and hence have a single point of failure. We are interested in decentralising view-overlap recognition  and enable direct interaction across the cameras. To  this end, we decouple feature extraction and cross-camera view-overlap recognition and define (and design) how cameras exchange features. Specifically, we decentralise an efficient two-stage approach that was originally devised for a single, moving camera~\cite{Galvez2012TRO} and design a framework to recognise view-overlaps across two hand-held or wearable cameras simultaneously moving in an unknown environment. Each camera independently extracts, at frame level, view and local features. A camera then recognises view overlaps by matching the features of a query view from another camera with the features of a previous views, and geometrically validates the matched view through the epipolar constraint. 

Our main contributions are a decentralised framework for cross-camera view-overlap recognition that decouples the extraction of the view features from the view-overlap recognition in support of the other camera, and is generic for different view and local features; and a dataset of image sequence pairs collected with hand-held and wearable cameras in four different environments, along with annotations of camera poses and view overlaps\footnote{The new sequences and all annotations can be found at: \url{https://www.eecs.qmul.ac.uk/~ax300/xview}}.

\section{Related Work}
\label{sec:relatedwork}

View-overlap recognition uses image-level features (or view features) and local features to represent each view. This process can be centralised or decentralised. A \textit{centralised} approach reconstructs, in a single processing unit, the 3D map using images or features received from all the cameras~\cite{Forster2013IROS,Riazuelo2014RAS,Schmuck2019JFR_CCMSLAM,Zou2013TPAMI}. With a \textit{decentralised} approach, cameras exchange local and view features~\cite{Cieslewski2018ICRA,cieslewski2017RAL}. 
Each camera can also create its own map and the cross-camera view matching can rely on a coarse-to-fine strategy~\cite{Cieslewski2018ICRA,Forster2013IROS,Schmuck2019JFR_CCMSLAM}, which first searches and matches the most similar view-feature in a database of accumulated view-features over time compared to the query view-feature and then matches local features associated with the 3D points in the local map. The coarse-to-fine strategy is important as view features are more sensitive to geometric differences, and hence additional steps, such as local feature matching and  geometric verification~\cite{Hartley2003MVG}, are often necessary to validate a matched view~\cite{Cieslewski2018ICRA,Galvez2012TRO,Garcia-Fidalgo2018RAL,MurArtal2015TRO,Schlegel2018RAL}. 

Local features can be aggregated into hierarchical structures, such as bag of visual words~\cite{Galvez2012TRO,Garcia-Fidalgo2018RAL,MurArtal2014ICRA,Nister2006CVPR,Sivic2003ICCV} or binary search tree~\cite{Schlegel2018RAL}, spatio-temporal representations~\cite{Tsintotas2019RAL,Xompero2020TIP_MST}, compact vectors summarising the first order statistics~\cite{Jegou2011TPAMI}, or Fisher vectors~\cite{Perronnin2010CVPR_FisherVectors}. Hierarchical structures can be pre-trained~\cite{Galvez2012TRO,MurArtal2014ICRA,Nister2006CVPR,Sivic2003ICCV} or built on-the-fly~\cite{Garcia-Fidalgo2018RAL,Schlegel2018RAL}, and can be used to speed up the search and matching of query local features with only a subset of local features stored in the structure. For example, Bags of Binary Words (DBoW)~\cite{Galvez2012TRO} coupled with binary features (e.g., Oriented FAST and Rotated BRIEF -- ORB~\cite{Rublee2011ICCV})  has been often preferred for 3D reconstruction and autonomous navigation due to the computational efficiency in feature extraction, searching, and matching, as well as the low storage requirements~\cite{Cieslewski2018ICRA,MurArtal2015TRO,Schmuck2019JFR_CCMSLAM}. However, as binary features are generally less robust to large geometric differences~\cite{Arandjelovic2018TPAMI,Balntas2018TPAMI,Lin2018TPAMI} than histogram-based features (e.g.,~SIFT\cite{Lowe2004IJCV}, RootSIFT~\cite{Arandjelovic2012CVPR}), centralised and decentralised approaches focused on the reconstruction aspect of the problem and on scenarios with substantial view-overlaps. 

Finally, no publicly available dataset exists with recordings from multiple cameras freely moving  in an environment with annotations of the camera poses and view overlaps. Because of the lack of publicly available datasets and their annotations, reproducibility and fair comparison of the previous methods are still an open challenge. Annotations are so far provided only for single camera sequences~\cite{Galvez2012TRO,Geiger2012CVPR,Schlegel2018RAL,Sturm2012IROS,Tsintotas2019RAL}. Zou and Tan~\cite{Zou2013TPAMI} released only one scenario with four camera sequences, but without annotations. To overcome the lack of datasets or their annotations, some  methods are evaluated on data acquired continuously with a single camera and then split into multiple parts to simulate a multi-camera scenario~\cite{Cieslewski2018ICRA,cieslewski2017RAL}. However, this results in highly overlapping portions, especially at the beginning and at the end of the sub-sequences. Alternatively, annotations of the camera poses and view overlaps can be determined through image-based 3D reconstruction methods~\cite{Schonberger2016CVPR}, but with the assumption that there are no moving objects or people in the scene. 

\section{Decentralising the Recognition of View Overlaps}
\label{sec:method}

\begin{figure}[t!]
  \centering
  \includegraphics[width=0.8\linewidth]{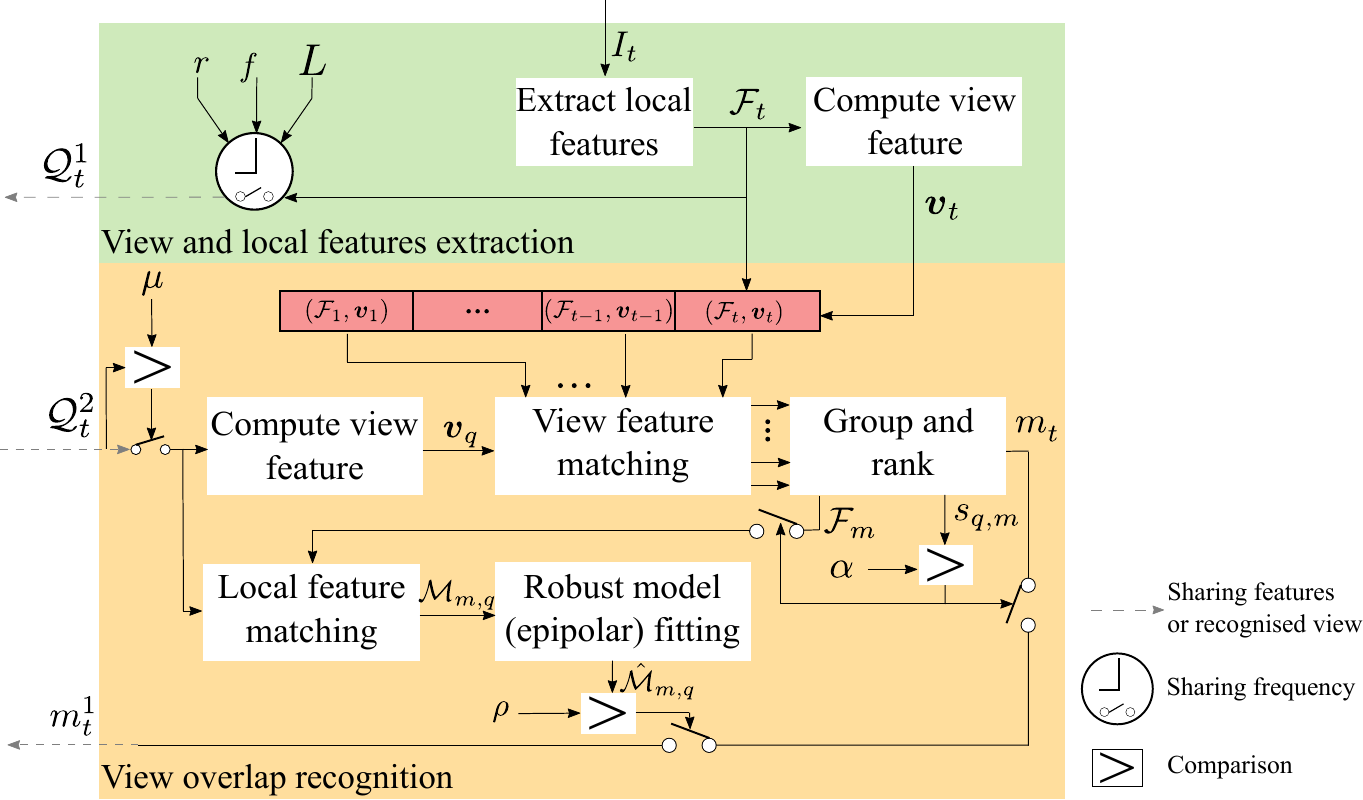}
%   \includesvg[inkscapelatex=false,width=0.8\linewidth]{blockdiagram}
  \caption{Block diagram of the proposed decentralised framework for cross-camera view-overlap recognition. For each camera, the framework decouples the extraction of view features $\boldsymbol{v}_t$ and local features $\mathcal{F}_t = \{ \mtrx{x}_{i,t}, \mtrx{d}_{i,t} \}_{i=1}^{F_t}$, for each frame $I_t$, from the recognition of view overlaps. While features are aggregated into a hierarchical structure over time (bag of binary words, shown as a red registry), each camera shares the set of local features $\mathcal{Q}_t$ at a pre-defined sharing rate $f$ and after an initialisation window of length $L$ (clock symbol). The camera receiving $\mathcal{Q}_t$ re-aggregates the received local features into the view feature $\boldsymbol{v}_q$ if their number is higher than $\mu$, and matches $\boldsymbol{v}_q$  with all the view features up to frame $t$. The local features of the matched view $m_t$, $\mathcal{F}_m$, whose score $s_{q,m}$ was the highest and higher than the threshold $\alpha$, are then matched with the query local features, obtaining $\mathcal{M}_{m,q}$. The matched view $m_t$ is validated through robust fitting of the epipolar geometry only if the number of inliers $|\hat{\mathcal{M}}_{m,q}| > \rho$. For keeping the notation simple, we use the camera indexes 1 and 2 only for denoting the shared query features and matched view.}
  \label{fig:blockdiagram}
%   \vspace{-10pt}
\end{figure}

In this section, we discuss our design choices (the distributed messaging, the features to extract and share, and the coarse-to-fine recognition strategy) to  decentralise DBoW~\cite{Galvez2012TRO}, an approach for real-time view-overlap recognition for a single moving camera that is based on binary features, corresponding bag of visual words, and a coarse-to-fine matching strategy.  Fig.~\ref{fig:blockdiagram} illustrates the proposed framework focusing on the processing and communication of a camera. 

\subsection{Camera Messaging}

We use the ZeroMQ\footnote{\url{http://zeromq.org/}} distributed messaging to decentralise the recognition across cameras, and multi-threading to decouple (i) the processing of the sequence and (ii) the view-overlap recognition in support of the other camera. These two operations are performed independently from each other within a single camera. 

To account for reliability, we adopt the ZeroMQ request-reply strategy: after sending a message containing the query view features, the camera waits for a reply from the other camera before processing the new frame. The reply includes a status denoting if a view overlap has been recognised and the corresponding frame in the sequence of the other camera. 
We assume that the cameras exchange query views on a communication channel that is always available (i.e.~the cameras remain within communication range), reliable (i.e.~all query views are received), and ideal (i.e.~no communication delays).

\subsection{Sharing Query Views}

For each frame $t$, each camera\footnote{For simplicity of notation, we do not use a camera index.} independently extracts a feature representing the image (view feature), $\boldsymbol{v} \in \mathbb{R}^G$, and a set of local features, $\mathcal{F}_t = \{ (\mtrx{d}_{i,t}, \mtrx{x}_{i,t}) \}_{i=1}^{F_t}$, that include their  location $\mtrx{x}_{i,t} \in \mathbb{R}^2$ (or interest point) and the corresponding binary descriptor $\mtrx{d}_{i,t} \in \{0,1\}^D$. The camera aims at extracting a pre-defined maximum target number of features $F$, but the number of interest points localised in the image, $F_t$, could be lower than $F$ due to the content (e.g., presence of corners) and appearance of the scene. We use ORB~\cite{Rublee2011ICCV} as local feature and the bag of visual words devised by DBoW for binary features as view feature~\cite{Galvez2012TRO}.
Specifically, this view feature stores the binary features in a vocabulary tree that is trained offline, using hierarchical k-means. Therefore, we add the extracted view and local features for each new frame to the vocabulary for speeding up the indexing, searching, and matching. 

While moving, cameras share the features of the view a time $t$, $\mathcal{Q}_t = \mathcal{F}_t$, to query previous views of the scene observed by the other camera. Note that a camera shares only the set of local features because the view feature can be directly recovered by the other camera as an aggregation of the local features using the bag of visual words with a pre-trained vocabulary. 
On the contrary, if we choose a view feature that can be obtained  with a different approach, e.g. convolutional neural networks~\cite{Arandjelovic2018TPAMI,Lin2018TPAMI}, then a camera should also share the view feature as part of the query.

Moreover, each camera shares the query features with a pre-defined rate $f$ that can be slower than the camera acquisition rate $r$ to avoid exchanging consecutive redundant information. Alternatively, automatic frame selection can be devised~\cite{MurArtal2015TRO} and features would be shared only at the selected frames. 
As the camera receiving the query view features will match these features with the features extracted in its previous frames to identify a view overlap, each camera avoids to share query features during the first $L$ frames (initialisation window). This allows a camera to add view and local features to the vocabulary tree to enable cross-camera view-overlap recognition. The sharing of the query features occurs only when 
\begin{equation}
    max(t-L,-1) - \left\lfloor \frac{max(t-L,-1)}{r}f \right\rfloor \frac{r}{f} = 0,
\end{equation}
where $\left\lfloor \cdot \right\rfloor$ is the floor operation. As an example, let the acquisition rate be $r=30$ fps, the sharing rate be $f=6$ fps, and the number of binary features, whose dimensionality is $D=256$ bits, be at  maximum $F=1,000$. By including the interest point (32$\times$2 bits) associated with each binary feature, each camera exchanges at most 240 kilobyte per second (kB/s). If a view feature extracted with a convolutional neural network, for example NetVLAD~\cite{Arandjelovic2018TPAMI} whose dimensionality is $G=131,072$ bits (4,096 elements), then each camera exchanges at maximum about 256 kB/s.

\subsection{View-Overlap Recognition}
\label{subsec:placerec}

To recognise a view overlap, the camera receiving  $\mathcal{Q}_t$ first searches and matches the query view feature $\boldsymbol{v}_q$ within its local vocabulary tree up to the current frame $t$ using the following score~\cite{Galvez2012TRO},
\begin{equation}
    s(\boldsymbol{v}_q,\boldsymbol{v}_k) = 1- \frac{1}{2}\left| \frac{\boldsymbol{v}_q}{|\boldsymbol{v}_q |} - \frac{\boldsymbol{v}_k}{|\boldsymbol{v}_k |} \right|,
\end{equation}
where $k=1, \ldots, t$.
Following the approach of DBoW~\cite{Galvez2012TRO}, the matching of the view features can identify up to $Nr$ candidates with respect to the query view feature, depending on the acquisition rate $r$ and a minimum score $\alpha$. Matched view features are thus grouped to prevent consecutive images competing with each other~\cite{Galvez2012TRO}. Each group contains a minimum number of matched view-features depending on the acquisition rate $r$. Matched view-features in the same group have a maximum difference (number of frames) of $\beta r$. The group and view with the highest score is selected for further validation. 

After identifying the candidate view $m_t$, local features of both views are matched via nearest neighbour search. To avoid ambiguities, we discard matches based on a threshold $\gamma$ for the Hamming distance between the descriptors of two local features, $H(\mtrx{d}_{i}^q, \mtrx{d}_{j}^m)$, and a threshold $\delta$ for the distance ratio between the closest and the second closest neighbour (Lowe's ratio test~\cite{Lowe2004IJCV})\footnote{The threshold for the Lowe's ratio test is usually set in the interval $[0.6,0.8]$. The threshold for the Hamming distance is usually chosen based on the typical separation of matching and non-matching feature distributions for binary features.}. The final set of matched local features is
\begin{equation}
    \mathcal{M}_{q,m} = \left\lbrace (\mtrx{d}_{i}^q, \mtrx{x}_{i}^q, \mtrx{d}_{j}^m, \mtrx{x}_{j}^m) | H(\mtrx{d}_{i}^q, \mtrx{d}_{j}^m) < \gamma, \frac{H(\mtrx{d}_{i}^q, \mtrx{d}_{j}^m)}{H(\mtrx{d}_{i}^q, \mtrx{d}_{l}^m)} < \delta \right\rbrace,
\end{equation}
where $\mtrx{d}_{i}^q$ is the $i$-th query binary feature from $\mathcal{Q}_t$, and $\mtrx{d}_{j}^m$ $\mtrx{d}_{l}^m$ are the closest and second closest neighbours found in the candidate view $m_t$, with $j \neq l$.

Matched local features are then geometrically validated through the epipolar constraint by estimating a matrix that encodes the undergoing rigid transformation (fundamental matrix)~\cite{Hartley2003MVG}: $(\mtrx{x}_i^q) \mtrx{F} (\mtrx{x}_j^m)^\top = 0$. As a set of eight correspondences between the two views is necessary to estimate the fundamental matrix $\mtrx{F}$, we discard the candidate view $m_t$ if $|\mathcal{M}_{q,m}| < \mu$ ($\mu=8$), where $| \cdot |$ is the cardinality of a set. On the contrary, when $|\mathcal{M}_{q,m}| > \mu$, we use robust fitting model with Random Sample Consensus~\cite{Fischler1981ACM} to obtain the hypothesis (fundamental matrix) with the highest number of inliers -- i.e., matched features satisfying the epipolar constraint with a maximum re-projection error, $\hat{\mathcal{M}}_{q,m,}$. To enable the searching of multiple solutions while sampling different correspondences, we require a minimum of $\rho$ correspondences that are also inliers for the estimated fundamental matrix, i.e., $|\hat{\mathcal{M}}_{q,m,}| > \rho$.

%%%%%%%%%%%%%%%%%%%%%%%%%%%%%%%%%%%%%%%%%%%%%%%%%%%%%%%%%%%%%%%%%%%%%%%%%%%%%%%%
\section{Validation}
\label{sec:validation}

\subsection{Strategies under Comparison}

We compare the proposed framework based on DBoW with five alternative choices for local features, view features, and  matching strategy. We refer to each version of the framework based on the main alternative component: D-DBoW, D-NetVLAD, D-DeepBit, D-RootSIFT, D-SuperPoint, and D-SuperGlue.

D-NetVLAD replaces the matching of view features based on bag of visual binary words with view features directly extracted with a convolutional neural network that in the last layer aggregates the first order statistics of mid-level convolutional features, i.e., the residuals in different parts of the descriptors space weighted by a soft assignment (NetVLAD)~\cite{Arandjelovic2018TPAMI}. D-NetVLAD aims to reproduce the decentralised approach in \cite{Cieslewski2018ICRA} but without the pre-clustering assignment across multiple cameras (more than two).
D-DeepBit also uses a convolutional neural network to represent an image with a view feature consisting of binary values (DeepBit)~\cite{Lin2018TPAMI}. The parameters of the model are learned in an unsupervised manner with a Siamese network to achieve invariance to different geometric transformations (e.g., rotation and scale), enforce minimal quantisation error between the real-value deep feature and the binary code to increase the descriptiveness (quantisation loss), and maximise  the information capacity (entropy) for each bin by evenly  distributing the binary code (even-distribution loss). Both variants of the framework preserves the sharing of ORB features for the coarse-to-fine strategy and the efficient matching of the query binary features through the reconstructed bag of visual words vector at the receiving camera, after identifying the most similar view feature (NetVLAD or DeepBit).
D-RootSIFT replaces the ORB features with transformed SIFT features via L1 normalisation, square-root, and L2 normalisation to exploit the equivalence between the Hellinger distance when comparing histograms and the Euclidean distance in the transformed feature space and hence reduce the sensitivity to smaller bin values (RootSIFT)~\cite{Arandjelovic2012CVPR}. 
D-SuperPoint also replaces the binary features with local features that are extracted with a two-branch convolutional neural network trained in a self-supervised manner to jointly localise and describe interest points directly on raw input images, assuming that the model is a homography (SuperPoint)~\cite{DeTone2018CVPRw_SuperPoint}. 
D-SuperGlue is coupled with SuperPoint features and replaces the nearest neighbour matching by performing context aggregation, matching, and filtering in a single end-to-end architecture that consists of a Graph Neural Network and an Optimal Matching layer~\cite{Sarlin2020CVPR_SuperGlue}.
For fair comparison and analysis of the impact of the chosen local features, D-RootSIFT, D-SuperPoint, and D-SuperGlue use NetVLAD as view feature. D-RootSIFT and D-SuperPoint use the nearest neighbour strategy with Lowe's ratio test to match the local features. 

As sampling strategy for robust fitting model of the fundamental matrix, we also compare the use of RANSAC~\cite{Fischler1981ACM}
against MAGSAC++~\cite{Barath2020CVPR_MAGSAC++}, a recent robust estimator that uses a quality scoring function while avoiding the inlier-outlier decision, and showed to be fast, robust, and more geometrically accurate. RANSAC is used within D-DBoW, D-NetVLAD and D-DeepBit and we therefore provide the corresponding alternatives with MAGSAC++, whereas D-RootSIFT, D-SuperPoint, and D-SuperGlue use directly MAGSAC++.

\subsection{Implementation Details}

The framework is implemented in C++ with OpenCV, DLib, and ZeroMQ libraries. For evaluation purposes and reproducibility, we implemented the framework in a synchronous way, and ZeroMQ request-reply messages are sent every frame to keep the cameras synchronised. Moreover, we allow the camera that finishes earlier to process its own sequence to remain active and recognise view overlaps until the other camera also terminates its own sequence.  
For all variants of the framework, we target to extract a maximum of $F=1,000$ local features (i.e., either ORB, RootSIFT, or SuperPoint). When matching binary features, we set the maximum Hamming distance to $\gamma=50$ as threshold to validate a match (for $D=256$~\cite{Calonder2010ECCV,MurArtal2015TRO}), while the threshold for Lowe's ratio test (or nearest neighbour distance ratio)~\cite{Lowe2004IJCV} is $\delta=0.6$. 
For NetVLAD, we use the best model (VGG-16 + NetVLAD + whitening), trained on Pitts30k~\cite{Torii2015TPAMI}, as suggested by the authors~\cite{Arandjelovic2018TPAMI}, whereas we use DeepBit 32-bit model trained on CIFAR10~\cite{Krizhevsky2009TR_Cifar} for DeepBit~\cite{Lin2018TPAMI}. Similarly to the decentralised approach in Cieslewski \textit{et al.}'s work~\cite{Cieslewski2018ICRA}, we consider the first 128 components of the NetVLAD descriptors and we use Euclidean distance to match descriptor, while we use the Hamming distance for DeepBit 32-bit model. For both NetVLAD and DeepBit, we extract the view features offline for each image of the sequences, and we provide the features as input to the framework. For the parameters of DBoW, SuperPoint, and SuperGlue, we used the values provided in the original implementations. We thus set $N=50$, $\alpha=0.03$, $\beta=3$, and the number of levels for direct index retrieval in the vocabulary tree to 2 for DBoW~\cite{Galvez2012TRO}.
D-DBOW, D-NetVLAD, and D-DeepBit uses the RANSAC implementation provided by DLib with a minimum number of inliers, $\rho=12$. For fairness in the evaluation, D-RootSIFT, D-SuperPoint, and D-Superglue use the matched views identified by D-NetVLAD to extract and match the local features for the validation of the view. Note that the models of SuperPoint and SuperGlue were implemented in Python with PyTorch, and hence D-RootSIFT, D-SuperPoint, and D-Superglue are not part of the implemented framework. For MAGSAC++, we use the implementation provided by OpenCV 4.5.5 with a minimum number of inliers, $\rho=15$. For both RANSAC and MAGSAC++, we set the maximum number of iterations to 500, the probability of success to 0.99, and the maximum re-projection error to 2 pixels.

%%%%%%%%%%%%%%%%%%%%%%%%%%%%
\subsection{Dataset and View-overlap Annotation}

We use sequence pairs from four scenarios: two scenarios that we collected with both hand-held and chest-mounted cameras -- \textit{gate} and \textit{backyard} of four sequences each -- and two publicly available datasets -- TUM-RGB-D~SLAM~\cite{Sturm2012IROS} and  \textit{courtyard}\footnote{\url{https://drone.sjtu.edu.cn/dpzou/dataset/CoSLAM/}} from CoSLAM~\cite{Zou2013TPAMI} -- for a total of $\sim$28,000 frames ($\sim$25~minutes). From TUM-RGB-D SLAM, we use the sequences fr1\_desk, fr1\_desk2, and fr1\_room to form the \textit{office} scenario. The \textit{courtyard} scenario consists of four sequences. We sub-sampled \textit{courtyard} from 50 to 25~fps and  \textit{backyard} from 30 to 10~fps for annotation purposes.
Tab.~\ref{tab:dataset} summarises the scenarios.

%%%%%%%%%%%%%%%%%%%%%%%%%%%%%%%%%%
\begin{table}[t!]
    \centering
    % \footnotesize
    \setlength\tabcolsep{4pt}
    \caption{Description of the scenarios with multiple hand-held moving cameras.
    % (\textit{backyard} has 2 wearable cameras).
    }
    % \vspace{-5pt}
    \begin{tabular}{lcrrrrcc}
    % \specialrule{1.2pt}{3pt}{1pt}
    \toprule
        \textbf{Scenario} & \textbf{\# Seq.} & \multicolumn{4}{c}{\textbf{Number of frames/seq.}} &  \textbf{fps} & \multicolumn{1}{c}{\textbf{Resolution}} \\
        % \specialrule{1.2pt}{0.2pt}{1pt}
        \toprule
       \textit{office} & 3 &  573 &  612 & 1,352 & --  & 30 & 640x480 \\
       \textit{courtyard} & 4 & 2,849 & 3,118 & 3,528 & 3,454 & 25 & 800x450 \\
      \textit{gate} & 4 & 330 & 450 & 480 & 375 & 30 & 1280x720 \\
      \textit{backyard} & 4 & 1,217 & 1,213 & 1,233 & 1,235 & 10 & 1280x720  \\
    % \specialrule{1.2pt}{0.2pt}{1pt}
    \bottomrule
    \multicolumn{8}{l}{\scriptsize{KEY~--~\# Seq.:~number of sequences; fps:~frame per second.}}
    \end{tabular}
    \label{tab:dataset}
    % \vspace{-12pt}
\end{table}
%%%%%%%%%%%%%%%%%%%%%%%%%%%%%%%%%%%%%%%%%%%%%%%%%%%%%%%%%%%%%%

\begin{figure}[t!]
    \centering
    \begin{tikzpicture}[
        pie chart,
        slice type={NFI}{dc1},
        slice type={bin1}{dc2},
        slice type={bin2}{dc3},
        slice type={bin3}{dc4},
        slice type={bin4}{dc5},
        slice type={bin5}{dc6},
        slice type={bin6}{dc7},
        slice type={bin7}{dc8},
        slice type={bin8}{dc9},
        slice type={bin9}{dc10},
        slice type={bin10}{dc11},
        pie values/.style={font={\tiny}},
        scale=1
    ]
    \pie{\textit{courtyard}}{42.23/NFI, 9.74/bin1, 9.74/bin2, 8.80/bin3, 7.97/bin4, 7.01/bin5, 3.62/bin6, 3.26/bin7, 2.77/bin8, 2.41/bin9, 2.47/bin10}
    \end{tikzpicture}
    \begin{tikzpicture}[
        pie chart,
        slice type={NFI}{dc1},
        slice type={bin1}{dc2},
        slice type={bin2}{dc3},
        slice type={bin3}{dc4},
        slice type={bin4}{dc5},
        slice type={bin5}{dc6},
        slice type={bin6}{dc7},
        slice type={bin7}{dc8},
        slice type={bin8}{dc9},
        slice type={bin9}{dc10},
        slice type={bin10}{dc11},
        pie values/.style={font={\tiny}},
        scale=1
    ]
    \pie{\textit{gate}}{5.04/NFI, 34.16/bin1, 30.09/bin2, 29.84/bin3, 0.86/bin4}
    \end{tikzpicture}
    \begin{tikzpicture}[
        pie chart,
        slice type={NFI}{dc1},
        slice type={bin1}{dc2},
        slice type={bin2}{dc3},
        slice type={bin3}{dc4},
        slice type={bin4}{dc5},
        slice type={bin5}{dc6},
        slice type={bin6}{dc7},
        slice type={bin7}{dc8},
        slice type={bin8}{dc9},
        slice type={bin9}{dc10},
        slice type={bin10}{dc11},
        pie values/.style={font={\tiny}},
        scale=1
    ]
    \pie{\textit{backyard}}{41.30/NFI, 5.79/bin1, 10.63/bin2, 9.22/bin3, 9.10/bin4, 6.37/bin5, 4.12/bin6, 3.41/bin7, 3.23/bin8, 3.31/bin9, 3.52/bin10}
    \end{tikzpicture}
    %     \begin{tikzpicture}[
    %     pie chart,
    %      slice type={NFI}{dc1},
    %     slice type={bin1}{dc2},
    %     slice type={bin2}{dc3},
    %     slice type={bin3}{dc4},
    %     slice type={bin4}{dc5},
    %     slice type={bin5}{dc6},
    %     slice type={bin6}{dc7},
    %     slice type={bin7}{dc8},
    %     slice type={bin8}{dc9},
    %     slice type={bin9}{dc10},
    %     slice type={bin10}{dc11},
    %     pie values/.style={font={\tiny}},
    %     scale=1
    % ]
    % \legend[shift={(-5.5cm,-3cm)}, font={\tiny}]{{NFI}/NFI} 
    % \legend[shift={(-4.4cm,-3cm)}, font={\tiny}]{{0-15$^\circ$}/bin1}
    % \legend[shift={(-3.3cm,-3cm)}, font={\tiny}]{{15-30$^\circ$}/bin2}
    % \legend[shift={(-2.4cm,-3cm)}, font={\tiny}]{{30-45$^\circ$}/bin3}
    % \legend[shift={(-1.1cm,-3cm)}, font={\tiny}]{{45-60$^\circ$}/bin4}
    % \legend[shift={(0cm,-3cm)}, font={\tiny}]{{60-75$^\circ$}/bin5}
    % \legend[shift={(1.1cm,-3cm)}, font={\tiny}]{{75-90$^\circ$}/bin6}
    % \legend[shift={(2.2cm,-3cm)}, font={\tiny}]{{90-105$^\circ$}/bin7}
    % \legend[shift={(3.3cm,-3cm)}, font={\tiny}]{{105-120$^\circ$}/bin8}
    % \legend[shift={(4.4cm,-3cm)}, font={\tiny}]{{120-135$^\circ$}/bin9}
    % \legend[shift={(5.5cm,-3cm)}, font={\tiny}]{{135-150$^\circ$}/bin10}
    % \end{tikzpicture}
    % \vspace{-5pt}
    \caption{
    Distribution of the angular distances in the outdoor scenarios. 
    % NFI:~no frustum intersection.
    % Distribution of the angular distance in the dataset: 
    % {\protect\raisebox{2pt}{\protect\tikz \protect\draw[green,line width=2] (0,0.5) -- (0.3,0.5);}}~\textit{courtyard},
    % {\protect\raisebox{2pt}{\protect\tikz \protect\draw[blue,line width=2] (0,0.5) -- (0.3,0.5);}}~\textit{gate},
    % {\protect\raisebox{2pt}{\protect\tikz \protect\draw[red,line width=2] (0,0.5) -- (0.3,0.5);}}~\textit{backyard}. KEY~--~Rel. freq.:~relative frequency; NFI:~no frustum intersection.
    Legend:
   \protect\tikz \protect\draw[dc1,fill=dc1] (0,0) rectangle (1.ex,1.ex);~no frustum intersection,
   \protect\tikz \protect\draw[dc2,fill=dc2] (0,0) rectangle (1.ex,1.ex);~0--15$^\circ$,
   \protect\tikz \protect\draw[dc3,fill=dc3] (0,0) rectangle (1.ex,1.ex);~15--30$^\circ$,
   \protect\tikz \protect\draw[dc4,fill=dc4] (0,0) rectangle (1.ex,1.ex);~30--45$^\circ$,
   \protect\tikz \protect\draw[dc5,fill=dc5] (0,0) rectangle (1.ex,1.ex);~45--60$^\circ$,
   \protect\tikz \protect\draw[dc6,fill=dc6] (0,0) rectangle (1.ex,1.ex);~60--75$^\circ$,
   \protect\tikz \protect\draw[dc7,fill=dc7] (0,0) rectangle (1.ex,1.ex);~75--90$^\circ$,
   \protect\tikz \protect\draw[dc8,fill=dc8] (0,0) rectangle (1.ex,1.ex);~90--105$^\circ$,
   \protect\tikz \protect\draw[dc9,fill=dc9] (0,0) rectangle (1.ex,1.ex);~105--120$^\circ$,
   \protect\tikz \protect\draw[dc10,fill=dc10] (0,0) rectangle (1.ex,1.ex);~120--135$^\circ$,
   \protect\tikz \protect\draw[dc11,fill=dc11] (0,0) rectangle (1.ex,1.ex);~135--150$^\circ$,
    }
    \label{fig:anngate}
    % \vspace{-10pt}
\end{figure}
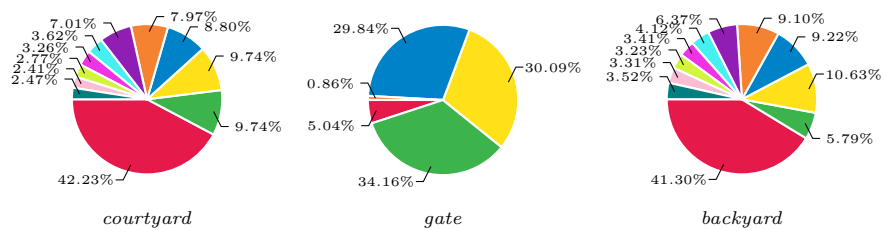

We automatically annotate correspondent views by exploiting camera poses and calibration parameters estimated with COLMAP~\cite{Schonberger2016CVPR}, a Structure-from-Motion pipeline, for \textit{gate}, \textit{courtyard}, and \textit{backyard}. These data are already available for \textit{office}. 
When the frustum between two views intersects under free space assumption~\cite{Moulon_OpenMVG}, we compute the viewpoint difference as the angular distance between the camera optical axes, the Euclidean distance between the camera positions, and the visual overlap as the ratio between the area spanned by projected 3D points within the image boundaries and the image area. Fig.~\ref{fig:anngate} visualises the distribution of the annotated angular distances across sequence pairs, showing how scenarios widely varies in their viewpoint differences based on the multiple camera motions. 
For \textit{office}, we localise a set of interest points (e.g.,~SIFT~\cite{Lowe2004IJCV}) in the first view and we back-project the points in 3D by exploiting the depth value at the corresponding location. For the other scenarios, we re-project the 3D points associated to the interest points of a frame in the first view onto the second view by using the estimated camera poses. When annotating correspondent views, we can define a threshold on the visual overlap, i.e.,~image pairs are a valid correspondence if their visual overlap is greater than the threshold. 
Note that a large overlap does not imply that the viewpoint is very similar as the angular and/or Euclidean distances can be large. Therefore, we constrain valid correspondent views to have an angular distance less than 70$^\circ$. In this work, we set the threshold on the overlap to 50\% and an analysis of its impact on the performance results will be subject of future work.
Moreover, we compute the total number of annotated views as the number of frames with at least one annotated view, i.e.,~a frame with more than one annotated view counts as one. 

\pgfplotstableread{binning_angular_distance.txt}\annbinang
\pgfplotstableread{binning_euclidean_distance.txt}\annbintran
\pgfplotstableread{binning_overlap_ratio.txt}\annboverlap

%%%%%%%%%%%%%%%%%%%%%%%%%%%%%%%%%%%%%%%%%%%%%%%

\subsection{Performance Measures}
\label{subsec:pereval}

We compare the framework with the different strategies using precision, recall, and accuracy as performance measures for each sequence pair in the dataset. We first compute the number of true positives (TP), false positives (FP), false negatives (FN), and true negatives (TN). A \textit{true positive} is a pair of views that is recognised as in correspondence (or overlap) and their annotation is also a valid correspondence. A \textit{false positive} is a pair of views that is recognised as in correspondence but their annotation is not. A \textit{false negative} is a (query) view with shared features from one camera that is not matched with any other view in the second camera (or vice versa) by a method, but a corresponding view is annotated. A \textit{true negative} is a view from one camera that is not matched with any other view in the second camera (or vice versa), and there is not any view that is annotated as correspondence in the second camera. 

\textit{Precision} is the number of correctly detected views over the total number of matched views: $P = TP/(TP + FP)$. \textit{Recall} is the number of correctly detected views in overlap over the total number of annotated view correspondences: $R = TP/(TP + FN)$. \textit{Accuracy} is the number of correctly recognised views in overlap and correctly rejected views over the total number of annotated view correspondences, correctly rejected views, and wrongly recognised views in overlap,
\begin{equation}
    A = \frac{TP + TN}{TP + FP + FN + TN}. 
    \label{eq:accuracy}
\end{equation}
Because of the pairwise approach, we sum TP, FP, TN, and FN, across the two cameras before computing P, R, and A. We will discuss the results using the performance measures as percentages with respect to each sequence pair.

\subsection{Parameter Setting}

We performed an analysis of the number of repetitions to perform due to the non-deterministic behaviour introduced by RANSAC, the length of the initialisation window $L$, the acquisition rate $r$ affecting the matching of view features, and the sharing rate $f$. We evaluated the accuracy of D-DBoW, D-NetVLAD, and D-DeepBit with 100 repetitions on two sequence pairs, \textit{gate 1$|$2} and \textit{office 1$|$2} ($X|Y$ denotes the indexes of the camera sequences in a scenario). We observed that the median accuracy was converging to a stable result within the first 30 repetitions due to the reproducibility of the same set of results for many repetitions. Therefore, the rest of the experiments are performed by repeating the framework on the dataset for 30 repetitions and reporting the median value. We also observed that the median accuracy is not significantly affected when varying the length of the initialisation window and the acquisition rate. We therefore set $L=30$ to allow the cameras to accumulate an initial set of view features, and $r$ to the corresponding frame rate of the scenario in Tab.~\ref{tab:dataset}. We set $f=6$ fps after observing that an automatic frame selection algorithm based of feature point tracking was re-initialising the feature trajectories every 5 frames, on average, for a handheld camera, while a person is walking.

SuperGlue provides pre-trained weights for both outdoor and indoor scenes. We use the outdoor pre-trained weights for the scenarios \textit{gate}, \textit{courtyard}, and \textit{backyard}, and the indoor pre-trained weights for \textit{office}.

%%%%%%%%%%%%%%%%%%%%%%%%%%%%%
\pgfplotstableread{global_vs_local_dbow.txt}\globalvslocal
\pgfplotstablegetrowsof{\globalvslocal}
\pgfmathtruncatemacro{\N}{\pgfplotsretval-1}
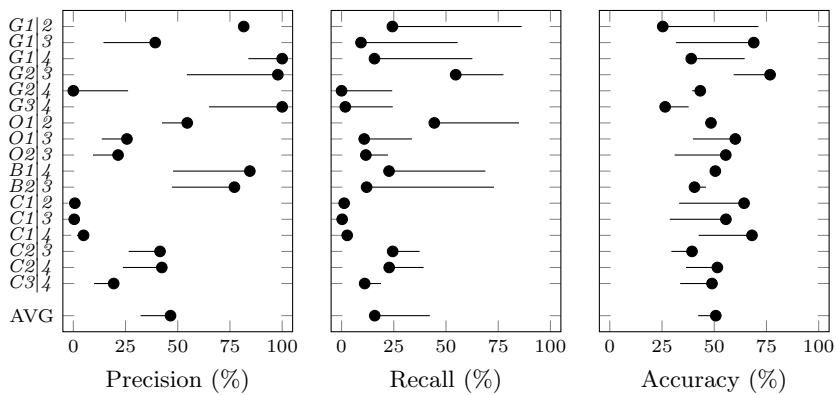
\begin{figure}[t!]
    \centering
    \begin{tikzpicture}
        \begin{axis}[
            y dir=reverse,
            width=0.38\columnwidth,
            height=0.48\columnwidth,
            xlabel={Precision (\%)},
            xmin=-5, xmax=105,
            xtick={0,25,50,75,100},
            ymin=0, ymax=20,
            xlabel near ticks,
            ylabel near ticks,
            ytick=data,
            yticklabels={\textit{G1$|$2},\textit{G1$|$3},\textit{G1$|$4},\textit{G2$|$3},\textit{G2$|$4},\textit{G3$|$4},\textit{O1$|$2},\textit{O1$|$3},\textit{O2$|$3},\textit{B1$|$4},\textit{B2$|$3},\textit{C1$|$2},\textit{C1$|$3},\textit{C1$|$4},\textit{C2$|$3},\textit{C2$|$4},\textit{C3$|$4}, AVG},
            label style={font=\footnotesize},
            tick label style={font=\scriptsize},
        ]
        \addplot+[only marks, color=white, mark=*, mark options={fill=white}] table[x=P-G, y=Scenario]{\globalvslocal}
        \foreach \i in {0,...,\N} {
            coordinate [pos=\i/\N] (a\i)
        }
        ;
        \addplot+[only marks, color=black, mark=*, mark options={fill=black}] table[x=P-GL, y=Scenario]{\globalvslocal}
        \foreach \i in {0,...,\N} {
            coordinate [pos=\i/\N] (b\i)
        }
        ;
        % \addplot+[only marks, color=red, mark=*, mark options={fill=red}] table[x=P-GL-N, y=Scenario]{\globalvslocal}
        % \foreach \i in {0,...,\N} {
        %     coordinate [pos=\i/\N] (c\i)
        % }
        % ;
        % \addplot[
        %     only marks,
        %     nodes near coords,
        %     % nodes near coords style={font=\tiny},
        %     every node near coord/.append style={xshift=23pt,yshift=0pt,anchor=east,font=\tiny},
        %     point meta=explicit symbolic,
        %     ] table [x=Dummy, y=Scenario, meta=P-C] {\globalvslocal};
        \end{axis}
        % now draw the connecting lines
        \foreach \i in {0,...,\N} {
            \draw (a\i) -- (b\i);
        }
        % \foreach \i in {0,...,\N} {
        %     \draw (a\i) -- (c\i);
        % }
    \end{tikzpicture}
    \begin{tikzpicture}
        \begin{axis}[
            y dir=reverse,
            width=0.38\columnwidth,
            height=0.48\columnwidth,
            xlabel={Recall (\%)},
            xmin=-5, xmax=105,
            xtick={0,25,50,75,100},
            ymin=0, ymax=20,
            xlabel near ticks,
            ylabel near ticks,
            ytick=data,
            yticklabels={},
            label style={font=\footnotesize},
            tick label style={font=\scriptsize},
        ]
        \addplot+[only marks, color=white, mark=*, mark options={fill=white}] table[x=R-G, y=Scenario]{\globalvslocal}
        \foreach \i in {0,...,\N} {
            coordinate [pos=\i/\N] (a\i)
        }
        ;
        \addplot+[only marks, color=black, mark=*, mark options={fill=black}] table[x=R-GL, y=Scenario]{\globalvslocal}
        \foreach \i in {0,...,\N} {
            coordinate [pos=\i/\N] (b\i)
        }
        ;
        % \addplot[
        %     only marks,
        %     nodes near coords,
        %     % nodes near coords style={font=\tiny},
        %     every node near coord/.append style={xshift=23pt,yshift=0pt,anchor=east,font=\tiny},
        %     point meta=explicit symbolic,
        %     ] table [x=Dummy, y=Scenario, meta=R-C] {\globalvslocal};
        \end{axis}
        % now draw the connecting lines
        \foreach \i in {0,...,\N} {
            \draw (a\i) -- (b\i);
        }
    \end{tikzpicture}
    \begin{tikzpicture}
        \begin{axis}[
            y dir=reverse,
            width=0.38\columnwidth,
            height=0.48\columnwidth,
            xlabel={Accuracy (\%)},
            xmin=-5, xmax=105,
            xtick={0,25,50,75,100},
            ymin=0, ymax=20,
            xlabel near ticks,
            ylabel near ticks,
            ytick=data,
            yticklabels={},
            label style={font=\footnotesize},
            tick label style={font=\scriptsize},
        ]
        \addplot+[only marks, color=white, mark=*, mark options={fill=white}] table[x=A-G, y=Scenario]{\globalvslocal}
        \foreach \i in {0,...,\N} {
            coordinate [pos=\i/\N] (a\i)
        }
        ;
        \addplot+[only marks, color=black, mark=*, mark options={fill=black}] table[x=A-GL, y=Scenario]{\globalvslocal}
        \foreach \i in {0,...,\N} {
            coordinate [pos=\i/\N] (b\i)
        }
        ;
        % \addplot[
        %     only marks,
        %     nodes near coords,
        %     % nodes near coords style={font=\tiny},
        %     every node near coord/.append style={xshift=23pt,yshift=0pt,anchor=east,font=\tiny},
        %     point meta=explicit symbolic,
        %     ] table [x=Dummy, y=Scenario, meta=A-C] {\globalvslocal};
        \end{axis}
        % now draw the connecting lines
        \foreach \i in {0,...,\N} {
            \draw (a\i) -- (b\i);
        }
    \end{tikzpicture}
    % \vspace{-7pt}
    \caption{Effect of matching both view and local features compared to matching only view features with D-DBoW. Positive change is shown with the black circles on the right of the line, whereas negative change with the black circles on the left. The length of the line denotes the magnitude of the change.
    KEY~--~sx$|$y:~image sequence pair where s is either G~(\textit{gate}), O~(\textit{office}), B~(\textit{backyard}), or C~(\textit{courtyard}), and x and y are the indexes of the first and second sequences in each scenario; AVG:~average.}
    \label{fig:globalvslocal}
    % \vspace{-12pt}
\end{figure}

\subsection{Discussion}

Fig.~\ref{fig:globalvslocal} shows the change in precision, recall, and accuracy from using only the first step (matching only view features) to using also the local feature matching and geometric validation in the 17 sequence pairs from the four scenarios in the dataset. The two-stage approach improves both precision and accuracy of, respectively, 14.35 percentage points (pp) and 8.43 pp, on average. This is observable in almost all sequence pairs as wrongly matched views are invalidated, while also increasing the recognition of views that are not in overlap (true negative in accuracy). However, the disadvantage of the two-stage approach is the large drop in recall (-26.37 pp, on average), as many views in overlap with respect to the query views were not identified.

%%%%%%%%%%%%%%%%%%%%%%%%%%%%%%%%%%%%%%%%%%5
% \FloatBarrier
\begin{table}[!]
    \centering
    \scriptsize
    \setlength\tabcolsep{5pt}
    \caption{Comparisons of precision (P), recall (R), and accuracy (A) using the framework with different strategies for each testing sequence pair.
    % \vspace{-10pt}
    }
    \begin{tabular}{lc|cc|c|rrrrrr}
    \toprule
    \textbf{Sequence} & \textbf{Pair} & \textbf{Q1} & \textbf{Q2} & \textbf{M} & 
    \textbf{D-DW} & \textbf{D-NV} & \textbf{D-DB} &  \textbf{D-RS} & \textbf{D-SP} & \textbf{D-SG} \\
    % \textbf{D-DBoW} & \textbf{D-NetVLAD} & \textbf{D-DeepBit} &  \textbf{D-RootSIFT} & \textbf{D-SuperPoint} & \textbf{D-SuperGlue} \\
    % \specialrule{1.2pt}{0.2pt}{1pt}
    \toprule
    \multirow{18}{*}{\textit{gate}} &
    \multirow{3}{*}{1$|$2} & \multirow{3}{*}{59} & \multirow{3}{*}{83} & P & 81.58 & \textbf{100.00} & \textbf{100.00} & 68.91 & 96.61 & 67.20 \\
    & & & & R & 24.33 & 11.76 & 11.76 & \textit{80.39} & 42.54 & \textbf{84.00} \\
    & & & & A & 25.35 & 15.49 & 15.49 & \textbf{59.86} & 44.37 & \textbf{59.86} \\
    \cmidrule(lr){2-11}
    & 
    \multirow{3}{*}{1$|$3} & \multirow{3}{*}{59} & \multirow{3}{*}{89} & P & 39.23 & \textit{40.00} & \textbf{53.57} & 29.51 & 0.00 & 26.92 \\
    & & & & R & 9.30 & 4.49 & 4.35 & \textit{43.90} & 0.00 & \textbf{70.00} \\
    & & & & A & \textit{68.92} & \textit{68.92} & \textbf{69.26} & 55.41 & 61.82 & 40.54 \\
    \cmidrule(lr){2-11}
    & 
    \multirow{3}{*}{1$|$4} & \multirow{3}{*}{59} & \multirow{3}{*}{68} & P & \textbf{100.00} & \textbf{100.00} & \textbf{100.00} & 90.91 & \textbf{100.00} & 75.00 \\
    & & & & R & 15.76 & 5.43 & 15.22 & \textit{54.35} & 31.52 & \textbf{60.00} \\
    & & & & A & 38.98 & 31.50 & 38.58 & \textbf{62.99} & 50.39 & \textit{59.84} \\
    \cmidrule(lr){2-11}
    & 
    \multirow{3}{*}{2$|$3} & \multirow{3}{*}{83} & \multirow{3}{*}{89} & P & \textit{97.92} & \textbf{100.00} & \textbf{100.00} & 87.30 & \textbf{100.00} & 56.14 \\
    & & & & R & 54.65 & 34.48 & 9.20 & \textit{65.88} & 45.40 & \textbf{78.05} \\
    & & & & A & \textit{76.74} & 66.86 & 54.07 & \textbf{77.91} & 72.38 & 60.47 \\
    \cmidrule(lr){2-11}
    & 
    \multirow{3}{*}{2$|$4} & \multirow{3}{*}{83} & \multirow{3}{*}{68}  & P & \textbf{100.00} & 0.00 & \textbf{100.00} & \textbf{100.00} & \textbf{100.00} & 35.90 \\
    & & & & R & 1.79 & 0.00 & 0.88 & \textit{11.50} & 4.42 & \textbf{14.74} \\
    & & & & A & 26.49 & 25.17 & 25.83 & \textbf{33.77} & 28.48 & \textit{29.80} \\
    \cmidrule(lr){2-11}
    &
    \multirow{3}{*}{3$|$4} & \multirow{3}{*}{89} & \multirow{3}{*}{68}  & P & 0.00 & 0.00 & 0.00 & 0.00 & 0.00 & 0.00 \\
    & & & & R & 0.00 & 0.00 & 0.00 & 0.00 & 0.00 & 0.00 \\
    & & & & A & 43.31 & \textbf{49.04} & \textit{47.77} & 36.94 & 46.82 & 29.94 \\
    \midrule
    \multirow{12}{*}{\textit{office}}
    & 
    \multirow{3}{*}{1$|$2} & \multirow{3}{*}{108} & \multirow{3}{*}{116} & P & \textbf{54.55} & 42.68 & 20.00 & \textit{44.07} & 42.86 & 29.25 \\
    & & & & R & \textit{44.40} & 10.75 & 0.66 & 20.23 & 13.24 & \textbf{53.75} \\
    & & & & A & \textbf{48.44} & 35.04 & 30.80 & \textit{39.29} & 36.61 & 37.05 \\
    \cmidrule(lr){2-11}
    & 
    \multirow{3}{*}{1$|$3} & \multirow{3}{*}{108} & \multirow{3}{*}{264} & P & \textit{25.68} & 16.67 & 0.00 & 20.59 & \textbf{28.22} & 17.31 \\
    & & & & R & \textit{10.84} & 0.70 & 0.00 & 5.51 & 4.83 & \textbf{20.45} \\
    & & & & A & 60.08 & 60.22 & \textit{60.48} & \textit{60.48} & \textbf{61.16} & 58.06 \\
    \cmidrule(lr){2-11}
    & 
    \multirow{3}{*}{2$|$3} & \multirow{3}{*}{116} & \multirow{3}{*}{264} & P & 21.43 & \textbf{35.15} & \textit{33.33} & 22.58 & 28.08 & 15.31 \\
    & & & & R & \textit{11.63} & 3.24 & 0.62 & 4.67 & 5.25 & \textbf{13.64} \\
    & & & & A & 55.53 & \textbf{57.37} & \textit{57.11} & 56.05 & 56.58 & 53.16 \\
    \midrule
    \multirow{6}{*}{\textit{backyard}}
    & 
    \multirow{3}{*}{1$|$4} & \multirow{3}{*}{237} & \multirow{3}{*}{240} & P & \textbf{84.52} & \textit{76.47} & 0.00 & 66.67 & 64.29 & 36.47 \\
    & & & & R & \textit{22.72} & 4.79 & 0.00 & 20.07 & 19.49 & \textbf{27.19} \\
    & & & & A & \textbf{50.52} & 40.67 & 38.16 & \textit{47.38} & 46.75 & 42.35 \\
    \cmidrule(lr){2-11}
    & 
    \multirow{3}{*}{2$|$3} & \multirow{3}{*}{236} & \multirow{3}{*}{240} & P & 77.14 & \textit{90.00} & \textbf{100.00} & 73.53 & 66.67 & 44.02 \\
    & & & & R & 11.99 & 3.34 & 0.32 & \textit{28.90} & 28.00 & \textbf{40.87} \\
    & & & & A & 40.55 & 35.92 & 34.03 & \textbf{47.90} & \textit{45.80} & 41.39 \\
    \midrule
    \multirow{18}{*}{\textit{courtyard}} 
    & 
    \multirow{3}{*}{1$|$2} & \multirow{3}{*}{553} & \multirow{3}{*}{613} & P & 0.73 & 0.00 & 0.00 & 0.00 & 0.00 & 0.60 \\
    & & & & R & 1.25 & 0.00 & 0.00 & 0.00 & 0.00 & 2.11 \\
    & & & & A & 64.28 & \textit{77.79} & \textbf{83.75} & 74.96 & 75.90 & 45.45 \\
    \cmidrule(lr){2-11}
    & 
    \multirow{3}{*}{1$|$3} & \multirow{3}{*}{553} & \multirow{3}{*}{693} & P & 0.45 & 0.00 & 0.00 & 0.68 & 0.00 & 0.62 \\
    & & & & R & 0.29 & 0.00 & 0.00 & 0.27 & 0.00 & 0.99 \\
    & & & & A & 55.58 & \textit{61.80} & \textbf{67.01} & 58.67 & 58.75 & 37.52 \\
    \cmidrule(lr){2-11}
    & 
    \multirow{3}{*}{1$|$4} & \multirow{3}{*}{553} & \multirow{3}{*}{678} & P & 4.91 & 5.22 & 0.00 & 4.60 & 3.86 & 2.13 \\
    & & & & R & 2.70 & 1.06 & 0.00 & 1.43 & 1.45 & 2.93 \\
    & & & & A & 68.07 & \textit{72.26} & \textbf{74.90} & 70.92 & 69.86 & 54.96 \\
    \cmidrule(lr){2-11}
    & 
    \multirow{3}{*}{2$|$3}  & \multirow{3}{*}{613} & \multirow{3}{*}{693} & P & \textbf{41.52} & \textit{40.34} & 21.32 & 39.32 & 37.92 & 19.65 \\
    & & & & R & \textbf{24.48} & 8.89 & 0.72 & 12.28 & 12.13 & \textit{21.67} \\
    & & & & A & \textbf{39.36} & 37.02 & 34.72 & \textit{38.82} & 38.17 & 27.45 \\
    \cmidrule(lr){2-11}
    & 
    \multirow{3}{*}{2$|$4} & \multirow{3}{*}{613} & \multirow{3}{*}{678} & P & 19.34 & \textbf{32.73} & 10.00 & \textit{31.43} & 30.43 & 12.58 \\
    & & & & R & \textit{11.06} & 4.86 & 0.35 & 6.08 & 6.48 & \textbf{14.53} \\
    & & & & A & 48.92 & \textbf{54.92} & 54.11 & \textbf{54.92} & \textit{54.69} & 41.36 \\
    \cmidrule(lr){2-11}
    & 
    \multirow{3}{*}{3$|$4} & \multirow{3}{*}{693} & \multirow{3}{*}{678} & P & \textit{42.38} & 38.83 & 29.79 & 41.18 & \textbf{47.95} & 23.32 \\
    & & & & R & \textit{22.81} & 5.65 & 0.81 & 9.21 & 11.97 & \textbf{26.10} \\
    & & & & A & \textbf{51.50} & 47.05 & 45.62 & 48.14 & \textit{49.53} & 42.01 \\
    \bottomrule
    \multicolumn{11}{l}{\scriptsize{\parbox{.99\linewidth}{QX:~number of queries from the first (second) camera; M:~performance measure; D-DW:~D-DBoW; D-NV:~D-NetVLAD; D-DB:~D-DeepBit; D-RS:~D-RootSIFT; D-SP:~D-SuperPoint; D-SG:~D-SuperGlue. For each row, the best- and second best-performing results are highlighted in \textbf{bold} and \textit{italic}, respectively. Best-performing results lower than 10 are not highlighted.}}}
    \end{tabular}
    \label{tab:resacc50}
\end{table}
% % %%%%%%%%%%%%%%%%%%%%%%%%%%%%%%%%%%%%%%%%%%%%%%%%%%%%%%%%%%%%%%%%

\pgfplotstableread{avgres.txt}\avgres
\pgfplotstableread{avgres_precision.txt}\avgresprec
\pgfplotstableread{avgres_recall.txt}\avgresrec
\begin{figure}[t!]
    \centering
    \begin{tikzpicture}
    \begin{axis}[
    axis lines*=box,
    ybar,
    width=\columnwidth,
    height=0.28\columnwidth,
	bar width=3.5pt,
    ymin=0,ymax=100,
    xtick=data,
    xticklabels={},
    tick label style={font=\footnotesize},
    % y=18.5mm,
    % y dir=reverse,
    xmin=0.5,  xmax=5.5,
    ylabel={Precision (\%)},
    label style={font=\footnotesize},
    % xmajorgrids=true,
    ]
    \addplot+[ybar, black, fill=c1, draw opacity=0.5, error bars/.cd, y dir=both, y explicit] table[x=Scenario, y=BW-m, y error =BW-s]{\avgresprec};
    \addplot+[ybar, black, fill=c2, draw opacity=0.5, error bars/.cd, y dir=both, y explicit] table[x=Scenario, y=NV-m, y error =NV-s]{\avgresprec};
    \addplot+[ybar, black, fill=c3, draw opacity=0.5, error bars/.cd, y dir=both, y explicit] table[x=Scenario, y=DB-m, y error =DB-s]{\avgresprec};
    \addplot+[ybar, black, fill=c4, draw opacity=0.5, error bars/.cd, y dir=both, y explicit] table[x=Scenario, y=BWM-m, y error =BWM-s]{\avgresprec};
    \addplot+[ybar, black, fill=c5, draw opacity=0.5, error bars/.cd, y dir=both, y explicit] table[x=Scenario, y=NVM-m, y error =NVM-s]{\avgresprec};
    \addplot+[ybar, black, fill=c6, draw opacity=0.5, error bars/.cd, y dir=both, y explicit] table[x=Scenario, y=DBM-m, y error =DBM-s]{\avgresprec};
    \addplot+[ybar, black, fill=c7, draw opacity=0.5, error bars/.cd, y dir=both, y explicit] table[x=Scenario, y=RS-m, y error =RS-s]{\avgresprec};
    \addplot+[ybar, black, fill=c8, draw opacity=0.5, error bars/.cd, y dir=both, y explicit] table[x=Scenario, y=SP-m, y error =SP-s]{\avgresprec};
    \addplot+[ybar, black, fill=c9, draw opacity=0.5, error bars/.cd, y dir=both, y explicit] table[x=Scenario, y=SG-m, y error =SG-s]{\avgresprec};
    \end{axis}
    \end{tikzpicture}
    \begin{tikzpicture}
    \begin{axis}[
    axis lines*=box,
    ybar,
    width=\columnwidth,
    height=0.28\columnwidth,
	bar width=3.5pt,
    ymin=0,ymax=100,
    xtick=data,
    xticklabels={},
    tick label style={font=\footnotesize},
    % y=18.5mm,
    % y dir=reverse,
    xmin=0.5,  xmax=5.5,
    ylabel={Recall (\%)},
    label style={font=\footnotesize},
    % xmajorgrids=true,
    ]
    \addplot+[ybar, black, fill=c1, draw opacity=0.5, error bars/.cd, y dir=both, y explicit] table[x=Scenario, y=BW-m, y error =BW-s]{\avgresrec};
    \addplot+[ybar, black, fill=c2, draw opacity=0.5, error bars/.cd, y dir=both, y explicit] table[x=Scenario, y=NV-m, y error =NV-s]{\avgresrec};
    \addplot+[ybar, black, fill=c3, draw opacity=0.5, error bars/.cd, y dir=both, y explicit] table[x=Scenario, y=DB-m, y error =DB-s]{\avgresrec};
    \addplot+[ybar, black, fill=c4, draw opacity=0.5, error bars/.cd, y dir=both, y explicit] table[x=Scenario, y=BWM-m, y error =BWM-s]{\avgresrec};
    \addplot+[ybar, black, fill=c5, draw opacity=0.5, error bars/.cd, y dir=both, y explicit] table[x=Scenario, y=NVM-m, y error =NVM-s]{\avgresrec};
    \addplot+[ybar, black, fill=c6, draw opacity=0.5, error bars/.cd, y dir=both, y explicit] table[x=Scenario, y=DBM-m, y error =DBM-s]{\avgresrec};
    \addplot+[ybar, black, fill=c7, draw opacity=0.5, error bars/.cd, y dir=both, y explicit] table[x=Scenario, y=RS-m, y error =RS-s]{\avgresrec};
    \addplot+[ybar, black, fill=c8, draw opacity=0.5, error bars/.cd, y dir=both, y explicit] table[x=Scenario, y=SP-m, y error =SP-s]{\avgresrec};
    \addplot+[ybar, black, fill=c9, draw opacity=0.5, error bars/.cd, y dir=both, y explicit] table[x=Scenario, y=SG-m, y error =SG-s]{\avgresrec};
    \end{axis}
    \end{tikzpicture}
    \begin{tikzpicture}
    \begin{axis}[
    axis lines*=box,
    ybar,
    width=\columnwidth,
    height=0.28\columnwidth,
	bar width=3.5pt,
    ymin=0,ymax=100,
    xtick=data,
    xticklabels={\textit{gate}, \textit{office}, \textit{backyard}, \textit{courtyard}, \textit{all scenarios}},
    tick label style={font=\footnotesize},
    % y=18.5mm,
    % y dir=reverse,
    xmin=0.5,  xmax=5.5,
    ylabel={Accuracy (\%)},
    label style={font=\footnotesize},
    % xmajorgrids=true,
    ]
    \addplot+[ybar, black, fill=c1, draw opacity=0.5, error bars/.cd, y dir=both, y explicit] table[x=Scenario, y=BW-m, y error =BW-s]{\avgres};
    \addplot+[ybar, black, fill=c2, draw opacity=0.5, error bars/.cd, y dir=both, y explicit] table[x=Scenario, y=NV-m, y error =NV-s]{\avgres};
    \addplot+[ybar, black, fill=c3, draw opacity=0.5, error bars/.cd, y dir=both, y explicit] table[x=Scenario, y=DB-m, y error =DB-s]{\avgres};
    \addplot+[ybar, black, fill=c4, draw opacity=0.5, error bars/.cd, y dir=both, y explicit] table[x=Scenario, y=BWM-m, y error =BWM-s]{\avgres};
    \addplot+[ybar, black, fill=c5, draw opacity=0.5, error bars/.cd, y dir=both, y explicit] table[x=Scenario, y=NVM-m, y error =NVM-s]{\avgres};
    \addplot+[ybar, black, fill=c6, draw opacity=0.5, error bars/.cd, y dir=both, y explicit] table[x=Scenario, y=DBM-m, y error =DBM-s]{\avgres};
    \addplot+[ybar, black, fill=c7, draw opacity=0.5, error bars/.cd, y dir=both, y explicit] table[x=Scenario, y=RS-m, y error =RS-s]{\avgres};
    \addplot+[ybar, black, fill=c8, draw opacity=0.5, error bars/.cd, y dir=both, y explicit] table[x=Scenario, y=SP-m, y error =SP-s]{\avgres};
    \addplot+[ybar, black, fill=c9, draw opacity=0.5, error bars/.cd, y dir=both, y explicit] table[x=Scenario, y=SG-m, y error =SG-s]{\avgres};
    \end{axis}
    \end{tikzpicture}
    % \vspace{-5pt}
   \caption{Comparison of mean and standard deviation of precision, recall, and accuracy between the different alternatives of the framework for cross-camera view recognition on the four scenarios and across all scenarios. 
%   Note that value of precision, recall, and accuracy for each sequence pair is the median across 30 repetitions due to the presence of RANSAC or MAGSAC++.
   Legend:
   \protect\tikz \protect\draw[c1,fill=c1] (0,0) rectangle (1.ex,1.ex);~D-DBoW,
   \protect\tikz \protect\draw[c2,fill=c2] (0,0) rectangle (1.ex,1.ex);~D-NetVLAD,
   \protect\tikz \protect\draw[c3,fill=c3] (0,0) rectangle (1.ex,1.ex);~D-DeepBit,
   \protect\tikz \protect\draw[c4,fill=c4] (0,0) rectangle (1.ex,1.ex);~D-DBoW with MAGSAC++,
   \protect\tikz \protect\draw[c5,fill=c5] (0,0) rectangle (1.ex,1.ex);~D-NetVLAD with MAGSAC++,
   \protect\tikz \protect\draw[c6,fill=c6] (0,0) rectangle (1.ex,1.ex);~D-DeepBit with MAGSAC++,
   \protect\tikz \protect\draw[c7,fill=c7] (0,0) rectangle (1.ex,1.ex);~D-RootSIFT,
   \protect\tikz \protect\draw[c8,fill=c8] (0,0) rectangle (1.ex,1.ex);~D-SuperPoint,
   \protect\tikz \protect\draw[c9,fill=c9] (0,0) rectangle (1.ex,1.ex);~D-SuperGlue.
   }
    \label{fig:bindesccomp}
    % \vspace{-10pt}
\end{figure}
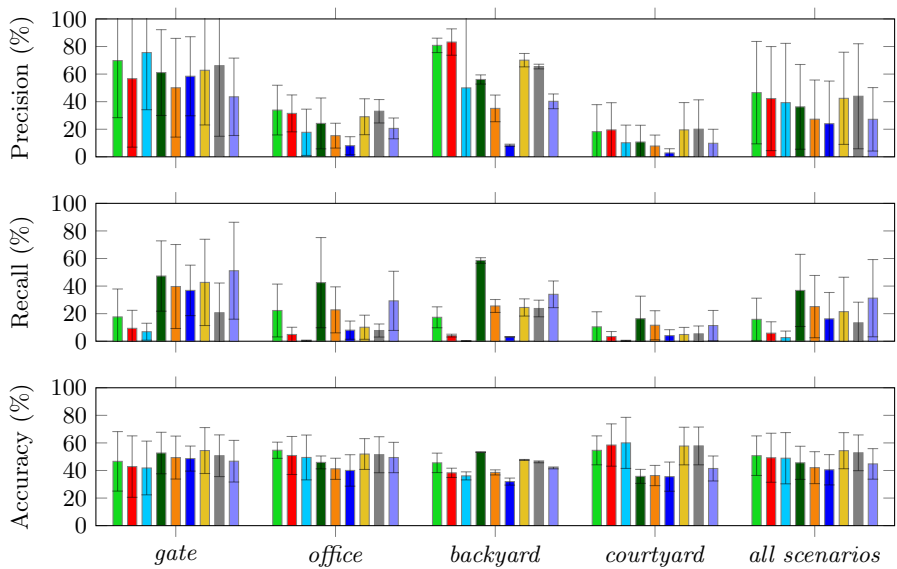
%%%%%%%%%%%%%%%%%%%%%%%%%%%%%%%%%%%%%%%%%%%%%%%%

Tab.~\ref{tab:resacc50} and Fig.~\ref{fig:bindesccomp} compare precision, recall, and accuracy of D-DBoW, D-NetVLAD, D-DeepBit, D-RootSIFT, D-SuperPoint, and D-SuperGlue. Overall, D-RootSIFT achieves the highest mean accuracy across all scenarios (54.38\%), followed by D-SuperPoint (52.83\%), and D-DBoW (50.74\%). Despite the 4 pp difference with D-RootSIFT, D-DBoW can benefit the fast inference of the bag of binary words and binary features for applications that may require nearly real-time processing. All variants have large standard deviations (around 10-20\% of the median accuracy), suggesting that different camera motions can affect the recognition of the view overlaps. Except for \textit{courtyard}, the bag of visual binary words of D-DBoW helps to achieve a higher accuracy than using D-NetVLAD and D-DeepBit as view features, independently of using RANSAC or MAGASC++ as robust estimator for validating the matched views. Note that MAGSAC++ does not contribute to improve the median accuracy of the cross-camera view-overlap recognition but achieves lower accuracy than RANSAC across all scenarios, except in some sequence pairs (e.g., in \textit{gate} and \textit{backyard}).
Tab~\ref{tab:resacc50} provides detailed results also in terms of precision and recall for each sequence pairs, including the number of query views shared by each camera. We can observe how different types of motion pairs can result in performance that can either accurately recognise the view overlaps and remove false positives or invalidating all query views despite being true correspondences (with the overlap threshold set at 50\%). There are challenging cases where the variants of the framework achieve a very low precision or recall, e.g., \textit{gate 3$|$4}, \textit{courtyard 1$|$2}, \textit{courtyard 1$|$3}, and \textit{courtyard 1$|$4}, but the accuracy is about 50\% or higher, showing that the frameworks can correctly recognise that there are no views in overlap for given query views. Overall, any of the variants of the framework recognises view overlaps that are not true correspondences and fails to identify many view overlaps in many sequences as we can see from median precision and recall that are lower than 50\% for most of the sequence pairs except for some of the scenario \textit{gate}.  Nevertheless, D-DBoW  is a competitive option for cross-camera view-overlap recognition with higher median precision on average when using RANSAC, and higher median recall on average when using MAGSAC++. 

\begin{figure}[tb!]
    \centering
    \setlength\tabcolsep{25pt}
    \begin{tabular}{ccc}
    D-DBoW & D-RootSIFT & D-SuperGlue \\
    \end{tabular}
    \\
    \includegraphics[width=0.27\columnwidth]{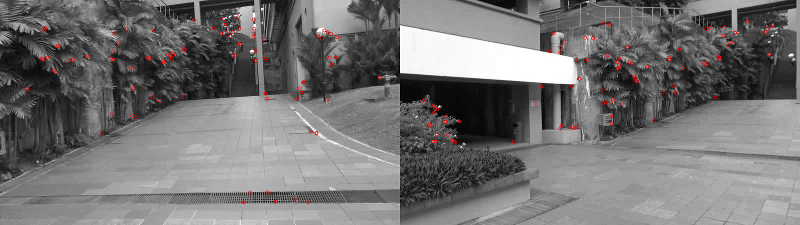}
    \includegraphics[width=0.27\columnwidth]{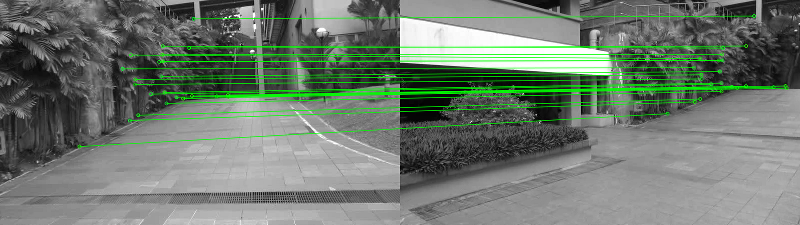}
    \includegraphics[width=0.27\columnwidth]{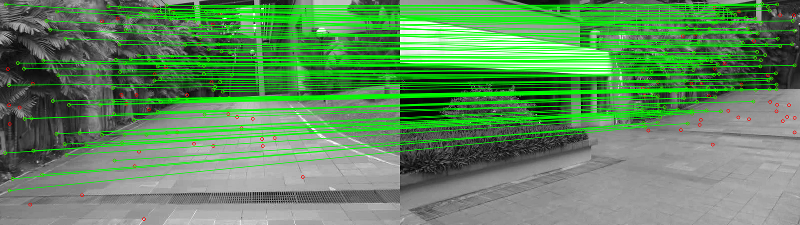}
    \includegraphics[width=0.27\columnwidth]{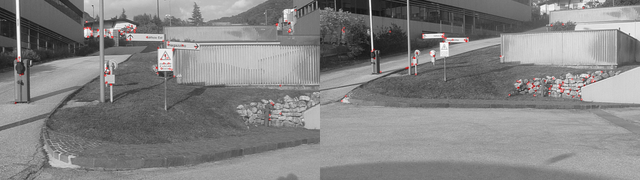}
    \includegraphics[width=0.27\columnwidth]{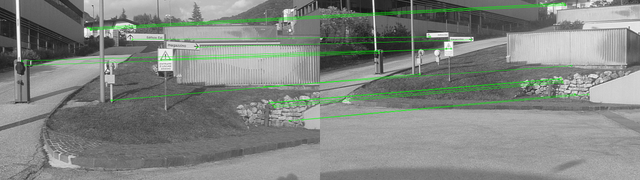}
    \includegraphics[width=0.27\columnwidth]{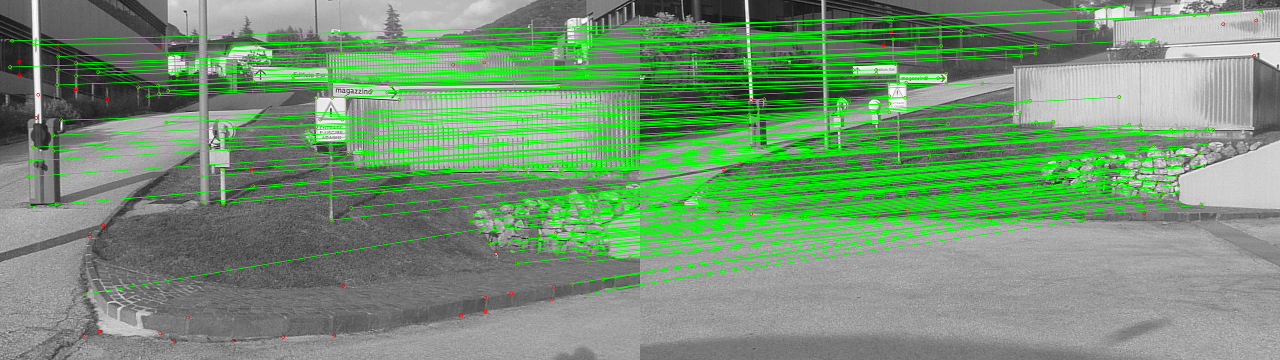}
    \includegraphics[width=0.27\columnwidth]{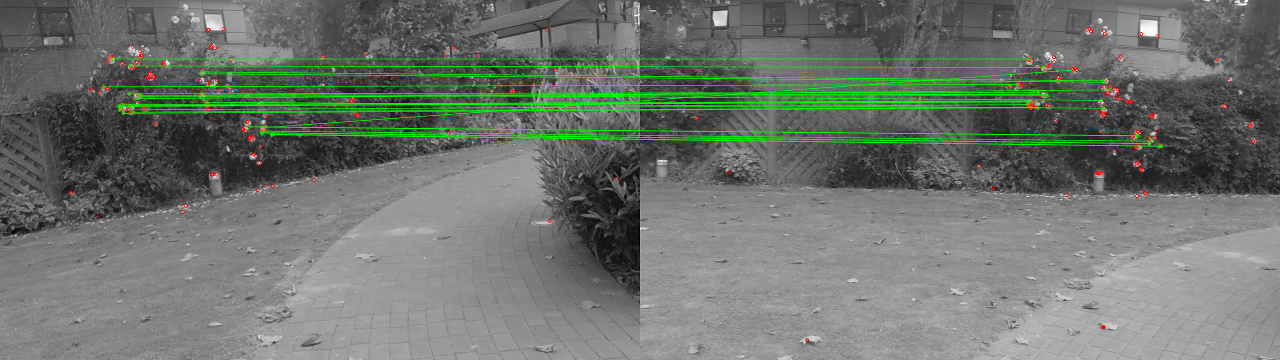}
    \includegraphics[width=0.27\columnwidth]{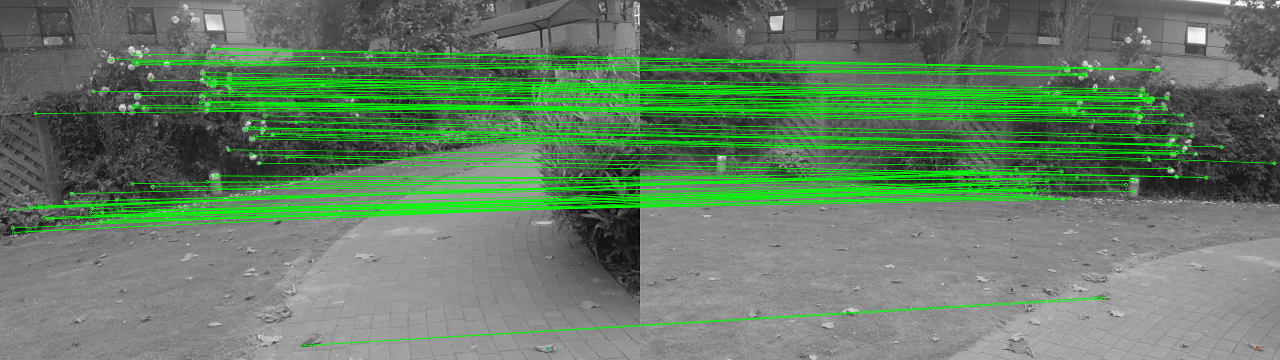}
    \includegraphics[width=0.27\columnwidth]{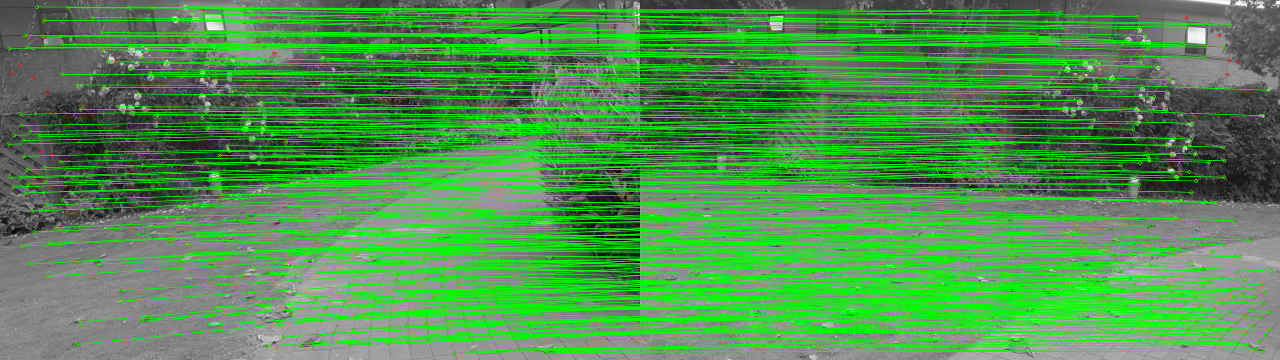}
    \includegraphics[width=0.27\columnwidth]{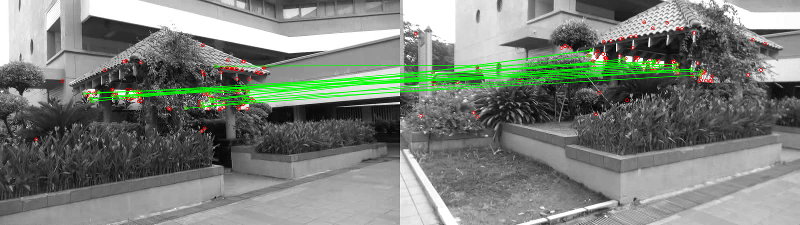}
    \includegraphics[width=0.27\columnwidth]{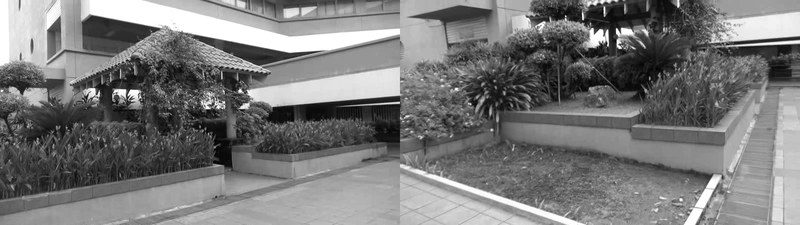}
    \includegraphics[width=0.27\columnwidth]{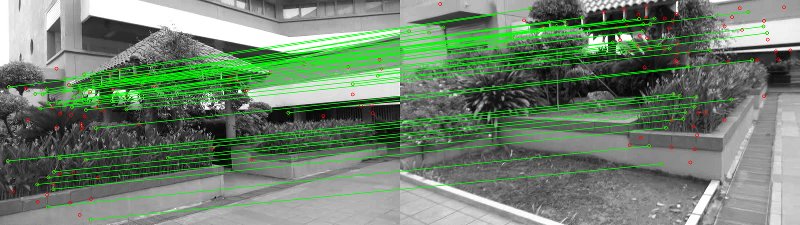}
    \includegraphics[width=0.27\columnwidth]{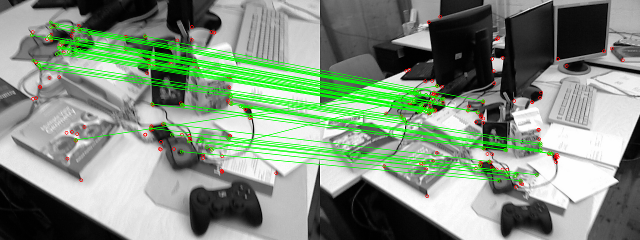}
    \includegraphics[width=0.27\columnwidth]{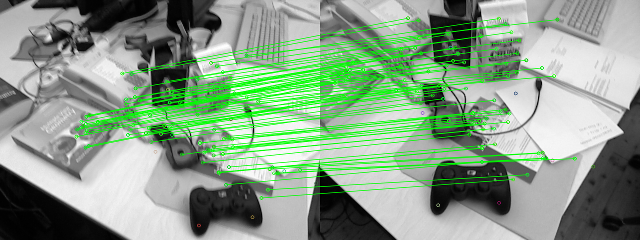}
    \includegraphics[width=0.27\columnwidth]{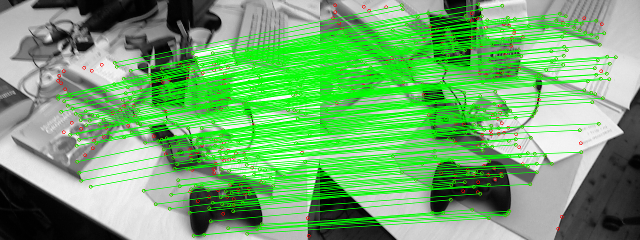}
    \includegraphics[width=0.27\columnwidth]{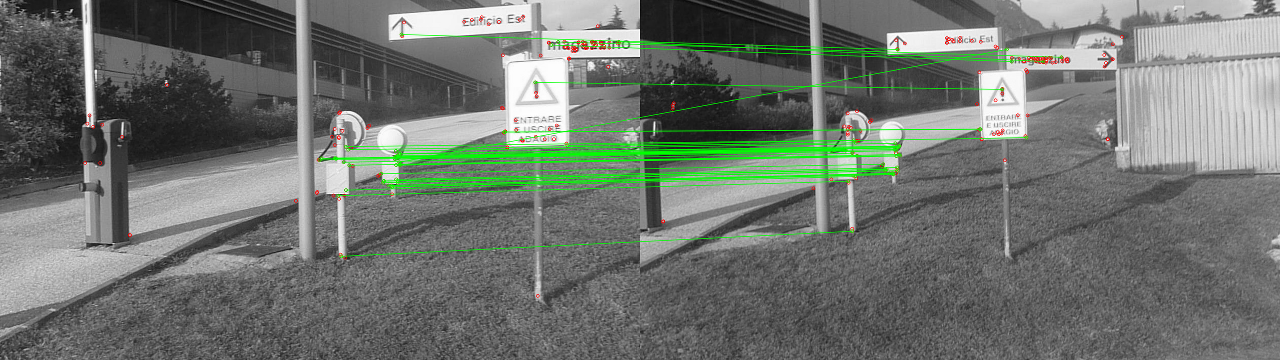}
    \includegraphics[width=0.27\columnwidth]{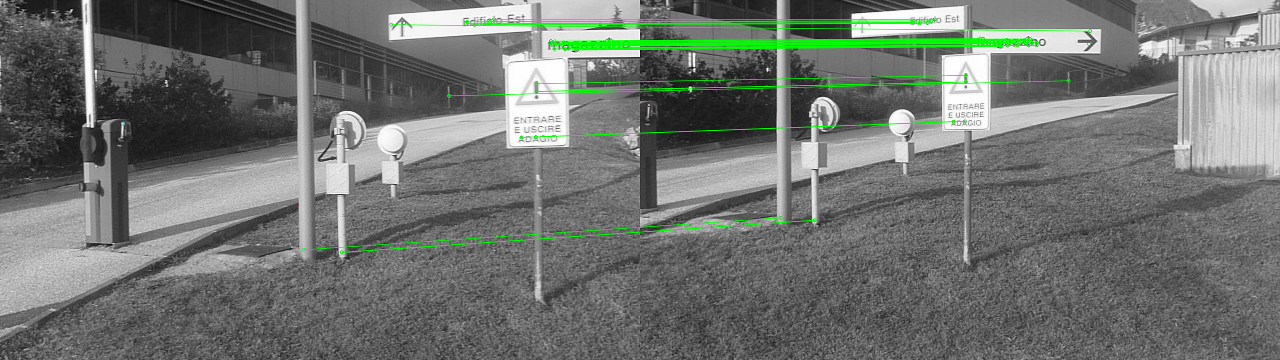}
    \includegraphics[width=0.27\columnwidth]{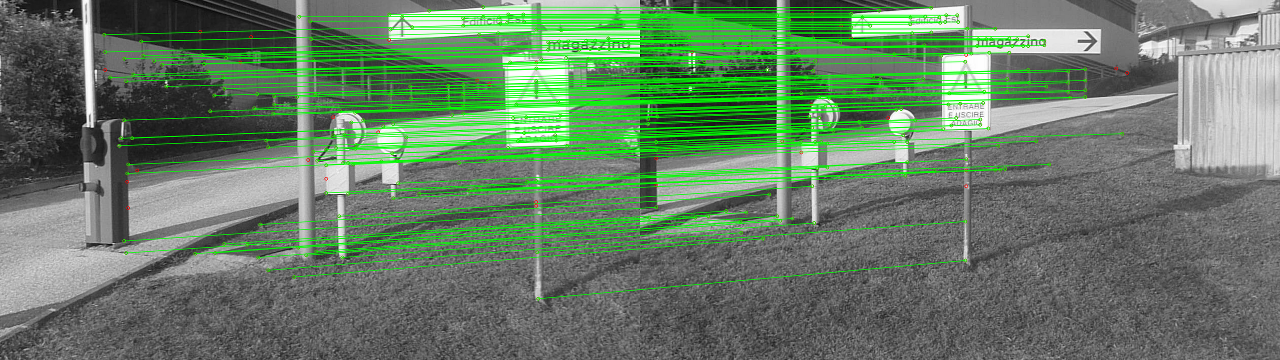}
    % \vspace{-5pt}
    \caption{Comparison of matched view features between D-DBoW, D-RootSIFT, and D-SuperGlue for selected query views from the sequence pairs C2$|$4, G1$|$3, B1$|$4, C2$|$3, O1$|$2, G2$|$3 (from top to bottom). Note that we show inliers (matched local features validated after RANSAC/MAGSAC++) as green lines and outliers as red circles. KEY~--~C:~\textit{courtyard}; B:~\textit{backyard}; G:~\textit{gate}; O:~\textit{office}.}
    \label{fig:vizmatches}
    % \vspace{-12pt}
\end{figure}

Fig.~\ref{fig:vizmatches} shows the matched views and matched local features (inliers after RANSAC or MAGSAC++) between D-DBoW, D-RootSIFT, and D-SuperGlue for sampled query views. As both D-RootSIFT and D-SuperGlue are using NetVLAD as view feature, the matched view is the same for both strategies, whereas D-DBoW can matched a different past views across cameras. Despite identifying a matched view, the framework can fail to match local features when the viewpoint or distance between the query and matched views increases (e.g., $>30^\circ$), showing the drop in recall discussed earlier. For example, we can observe this situation for D-DBoW and D-RootSIFT in the first two rows and the fourth row. 
On the contrary and as expected, the number of matched local features increases as the viewpoint is more similar between the query and the matched view (e.g., $<15^\circ$) if view overlaps are recognised. Moreover, the framework, independently of the strategy, tends to match local features in flat and localised areas in the outdoor scenes \textit{gate}, \textit{courtyard}, and \textit{backyard}, and on the corner of different objects in the \textit{office} scenario. However, if $|\mathcal{M}_{m,q}| < \rho$ or the epipolar constraint is not satisfied (e.g, all the local features are co-planar), then the matched view $m$ is not validated.
D-SuperGlue shows to match a higher number of local features compared to D-DBoW and D-RootSIFT, but it can also match wrong local features, as shown in \textit{courtyard} (fourth row).

\section{Conclusion}

We presented a decentralised framework to recognise overlapping views across freely moving cameras. The framework is modular and can be combined with various view and local features, and does not require a 3D map or any prior information about the environment.  
Through coarse-to-fine matching and  geometric validation, the framework  can recognise overlaps in past views of the other camera and  improve the average accuracy  by discarding wrongly matched views.
As future work, we will consider the scalability of the framework with more than two cameras and robustness with a lossy communication channel.

% \clearpage\mbox{}Page \thepage\ of the manuscript.
% \clearpage\mbox{}Page \thepage\ of the manuscript.

% \clearpage
\bibliographystyle{splncs04}
\bibliography{main}

\end{document}